# The Optimal Design of Three Degree-of-Freedom Parallel Mechanisms for Machining Applications

Damien Chablat - Philippe Wenger – Félix Majou
Institut de Recherche en Communications et Cybernétique de Nantes[1]
1, rue de la Noë, 44321 Nantes, France
Damien.Chablat@irccyn.ec-nantes.fr

## Abstract

The subject of this paper is the optimal design of a parallel mechanism intended for three-axis machining applications. Parallel mechanisms are interesting alternative designs in this context but most of them are designed for three- or six-axis machining applications. In the last case, the position and the orientation of the tool are coupled and the shape of the workspace is complex. The aim of this paper is to use a simple parallel mechanism with two-degree-of-freedom (dof) for translational motions and to add one leg to have one-dof rotational motion. The kinematics and singular configurations are studied as well as an optimization method. The three-degree-of-freedom mechanisms analyzed in this paper can be extended to four-axis machines by adding a fourth axis in series with the first two.

## 1  Introduction

Parallel kinematic machines (PKM) are commonly claimed to offer several advantages over their serial counterparts, like high structural rigidity, high dynamic capacities and high accuracy [1]. Thus, PKM are interesting alternative designs for high-speed machining applications.

The first industrial application of PKMs was the Gough platform, designed in 1957 to test tyres [2]. PKMs have then been used for many years in flight simulators and robotic applications [3] because of their low moving mass and high dynamic performances [1]. This is why parallel kinematic machine tools attract the interest of most researchers and companies. Since the first prototype presented in 1994 during the IMTS in Chicago by Gidding&Lewis (the VARIAX), many other prototypes have appeared.

To design a parallel mechanism, two important problems must be solved. The first one is the identification of singular configurations, which can be located inside the workspace. For a six-dof parallel mechanism, like the Gough-Stewart platform, the location of the singular configurations is very difficult to characterize and can change under small variations in the design parameters [3]. The second problem is the non-homogeneity of the performance indices (condition number, stiffness...) throughout the workspace [1]. To the authors' knowledge, only one parallel mechanism is isotropic throughout the workspace [4] but the legs are subject to bending. Moreover, this concept is limited to three-dof mechanisms and cannot be extended to four or five-dof parallel mechanisms.

Numerous papers deal with the design of parallel mechanisms [4,5]. However, there is a lack of four- or five-dof parallel mechanisms, which are especially required for machining applications [6].

To decrease the cost of industrialization of new PKM and to reduce the problems of design, a modular strategy can be applied. The translational and rotational motions can be divided into two separated parts to produce a mechanism where the direct kinematic problem is decoupled. This simplification yields also some simplifications in the definition of the singular configurations.

The organization of this paper is as follows. Next section presents design problems of parallel mechanisms. The kinematic description and singularity analysis of the parallel mechanism used, are reported in sections 3.1 and 3.2. Sections 3.3 and 3.4 are devoted to design and the optimization.

---

## 2 About parallel kinematic machines

### 2.1 General remarks

In a PKM, the tool is connected to the base through several kinematic chains or legs that are mounted in parallel. The legs are generally made of telescopic struts with fixed foot points (Figure 1a), or fixed length struts with moveable foot points (Figure 1b).

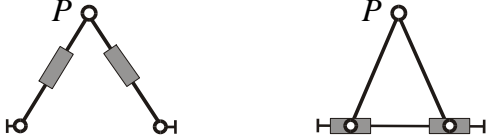

Figure 1a: A bipod PKM    Figure 1b: A biglide PKM

For machining applications, the second architecture is more appropriate because the masses in motion are lower. The linear joints can be actuated by means of linear motors or by conventional rotary motors with ball screws.
A classification of the legs suitable to produce motions for parallel kinematic machines is provided by [6] with their degrees of freedom and constraints. The connection of identical or different kinematic legs permits the authors to define two-, three-, four- and five-dof parallel mechanisms. However, it is not possible to remove one leg from a four-dof to produce a three-dof mechanism because no modular approach is used.

### 2.2 Singularities

The singular configurations (also called singularities) of a PKM may appear inside the workspace or at its boundaries. There are two types of singularities [7]. A configuration where a finite tool velocity requires infinite joint rates is called a serial singularity. These configurations are located at the boundary of the workspace. A configuration where the tool cannot resist any effort and in turn, becomes uncontrollable is called a parallel singularity. Parallel singularities are particularly undesirable because they induce the following problems (i) a high increase in forces in joints and links, that may damage the structure, and (ii) a decrease of the mechanism stiffness that can lead to uncontrolled motions of the tool though actuated joints are locked.

Figures 2a and 2b show the singularities for the biglide mechanism of Fig. 1b. In Fig. 2a, we have a serial singularity. The velocity amplification factor along the vertical direction is null and the force amplification factor is infinite.

Figure 2b shows a parallel singularity. The velocity amplification factor is infinite along the vertical direction and the force amplification factor is close to zero. Note that a high velocity amplification factor is not necessarily desirable because the actuator encoder resolution is amplified and thus the accuracy is lower.

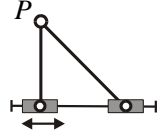 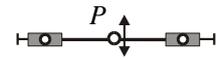

Figure 2a: A serial singularity    Figure 2b: A parallel singularity

The determination of the singular configurations for two-dof mechanisms is very simple; conversely, for a six-dof mechanism like the Gough-Stewart platform, a mechanism with six-dof, the problem is very difficult [3]. With a modular architecture, when the position and the orientation of the mobile platform are decoupled, the determination of the singularities is easier.

### 2.3 Kinetostatic performance of parallel mechanism

Various performance indices have been devised to assess the kinetostatic performances of serial and parallel mechanisms. The literature on performance indices is extremely rich to fit in the limits of this paper (service angle, dexterous workspace and manipulability…) [9]. The main problem of these performance indices is that they do not take into account the location of the tool frame. However, the Jacobian determinant depends on this location [9] and this location depends on the tool used.
To the authors' knowledge there is no parallel mechanism, suitable for machining, for which the kinetostatic performance indices are constant throughout the workspace (like the condition number or the stiffness…). For a serial three-axis machine tool, a motion of an actuated joint yields the same motion of the tool (the transmission factors are equal to one). For a parallel machine, these motions are generally not equivalent. When the mechanism is close to a parallel singularity, a small joint rate can generate a large velocity of the tool. This means that the positioning accuracy of the tool is lower in some directions for some configurations close to parallel singularities because the encoder resolution is amplified. In addition, a high velocity amplification factor in one direction is equivalent to a loss of stiffness in this direction. The manipulability ellipsoid of the Jacobian matrix of robotic manipulators was defined two decades ago [8]. The $\mathbf{JJ}^{-1}$ eigenvalues square roots, $\gamma_1$ and $\gamma_2$, are the lengths of the semi-axes of the ellipse that define the two velocity amplification factors between the actuated joints velocities and the velocity vector $\dot{\mathbf{t}}$ ($\lambda_1 = 1/\gamma_1$ and $\lambda_2 = 1/\gamma_2$). For parallel mechanisms with only pure translation or pure rotation motions, the variations of these factors inside the Cartesian workspace can be limited by the following constraints

$$\lambda_{i\,min} \leq \lambda_i \leq \lambda_{i\,max}$$

Unfortunately, this concept is quite difficult to apply when the tool frame can produce both rotational and translational motions. In this case, indeed the Jacobian matrix is not homogeneous [9].

A first way to solve this problem is its normalization by computing its characteristic length [9-10]. The second approach is to limit the values of all terms of the Jacobian matrix to avoid singular configuration and to associate these values to a physical measurement (See section 3.4).

## 3 Kinematics of mechanisms studied

### 3.1 Kinematics of a parallel mechanism for translational motions

The aim of this section is to define the kinematics and the singular configuration of a two-dof translational mechanism (Figure 3), which can be extended to three-axis machines by adding a third axis in series with the first two. The output body is connected to the linear joints through a set of two parallelograms of equal lengths $L = A_i B_i$, so that it can move only in translation.

The two legs are $PPa$ identical chains, where $P$ and $Pa$ stand for Prismatic and Parallelogram joints, respectively. This mechanism can be optimized to have a workspace whose shape is close to a square workspace and the velocity amplification factors are bounded [11].

The joint variables $\rho_1$ and $\rho_2$ are associated with the two prismatic joints. The output variables are the Cartesian coordinates of the tool center point $P = [x\ y]^T$. To control the orientation of the reference frame attached to $P$, two parallelograms can be used, which also increase the rigidity of the structure, Figure 3.

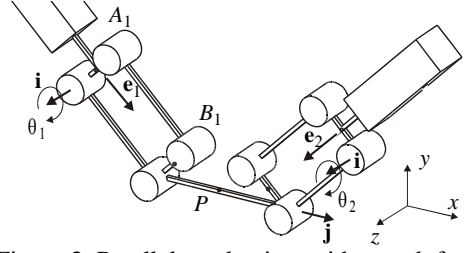

Figure 3: Parallel mechanism with two-dof

To produce the third translational motion, it is possible to place orthogonally a third prismatic joint.

The velocity $\dot{\mathbf{p}}$ of point $P$ can be expressed in two different ways. By traversing the closed loop $(A_1B_1P - A_2B_2P)$ in two possible directions, we obtain

$$\dot{\mathbf{p}} = \dot{\mathbf{a}}_1 + \dot{\theta}_1 \mathbf{i} \times (\mathbf{b}_1 - \mathbf{a}_1) \quad (1a)$$
$$\dot{\mathbf{p}} = \dot{\mathbf{a}}_2 + \dot{\theta}_2 \mathbf{i} \times (\mathbf{b}_2 - \mathbf{a}_2) \quad (1b)$$

where $\mathbf{a}_1$, $\mathbf{b}_1$, $\mathbf{a}_2$ and $\mathbf{b}_2$ represent the position vectors of the points $A_1$, $B_1$, $A_2$ and $B_2$, respectively. Moreover, the velocities $\dot{\mathbf{a}}_1$ and $\dot{\mathbf{a}}_2$ of $A_1$ and $A_2$ are given by $\dot{\mathbf{a}}_1 = \mathbf{e}_1 \dot{\rho}_1$ and $\dot{\mathbf{a}}_2 = \mathbf{e}_2 \dot{\rho}_2$, respectively.

For an isotropic configuration to exist where the velocity amplification factors are equal to one, we must have $\mathbf{e}_1 . \mathbf{e}_2 = 0$ [11] (Figure 5).

We would like to eliminate the two passive joint rates $\dot{\theta}_1$ and $\dot{\theta}_2$ from Eqs. (1a-b), which we do upon dot-multiply the former by $(\mathbf{b}_1 - \mathbf{a}_1)^T$ and the latter by $(\mathbf{b}_2 - \mathbf{a}_2)^T$, thus obtaining

$$(\mathbf{b}_1 - \mathbf{a}_1)^T \dot{\mathbf{p}} = (\mathbf{b}_1 - \mathbf{a}_1)^T \mathbf{e}_1 \dot{\rho}_1 \quad (2a)$$
$$(\mathbf{b}_2 - \mathbf{a}_2)^T \dot{\mathbf{p}} = (\mathbf{b}_2 - \mathbf{a}_2)^T \mathbf{e}_2 \dot{\rho}_2 \quad (2b)$$

Equations (2a-b) can be cast in vector form, namely $\mathbf{A}\dot{\mathbf{p}} = \mathbf{B}\dot{\boldsymbol{\rho}}$, with $\mathbf{A}$ and $\mathbf{B}$ denoted, respectively, as the parallel and serial Jacobian matrices,

$$\mathbf{A} \equiv \begin{bmatrix} (\mathbf{b}_1 - \mathbf{a}_1)^T \\ (\mathbf{b}_2 - \mathbf{a}_2)^T \end{bmatrix} \quad \mathbf{B} \equiv \begin{bmatrix} (\mathbf{b}_1 - \mathbf{a}_1)^T \mathbf{e}_1 & 0 \\ 0 & (\mathbf{b}_2 - \mathbf{a}_2)^T \mathbf{e}_2 \end{bmatrix}$$

where $\dot{\boldsymbol{\rho}}$ is defined as the vector of actuated joint rates and $\dot{\mathbf{p}}$ is the velocity of point $P$, i.e.,

$$\dot{\boldsymbol{\rho}} = \begin{bmatrix} \dot{\rho}_1 & \dot{\rho}_2 \end{bmatrix}^T \text{ and } \dot{\mathbf{p}} = \begin{bmatrix} \dot{x} & \dot{y} \end{bmatrix}^T$$

When $\mathbf{A}$ and $\mathbf{B}$ are not singular, we obtain the relations,

$$\dot{\mathbf{p}} = \mathbf{J}\dot{\boldsymbol{\rho}} \text{ with } \mathbf{J} = \mathbf{A}^{-1}\mathbf{B}$$

Parallel singularities occur whenever the lines $A_1B_1$ and $A_2B_2$ are colinear, i.e. when $\theta_1 - \theta_2 = k\pi$, for $k = 1,2,....$ Serial singularities occur whenever $\mathbf{e}_1 \perp \mathbf{b}_1 - \mathbf{a}_1$ or $\mathbf{e}_2 \perp \mathbf{b}_2 - \mathbf{a}_2$. To avoid these two singularities, the range limits are defined in using suitable bounds on the velocity factor amplification (See section 3.3).

## 3.2   Kinematics of a spatial parallel mechanism with one-dof of rotation

The aim of this section is to define the kinematics of a simple mechanism with two-dof of translation and one-dof of rotation. To be modular, the direct kinematic problem must be decoupled between position and orientation equations. A decoupled version of the Gough-Stewart Platform exists but it is very difficult to build because three spherical joints must coincide [12]. Thus, it cannot be used to perform milling applications. The main idea of the proposed architecture is to attach a new body with the tool frame to the mobile platform of the two-dof mechanism defined in the previous section. The new joint admits one or two-dofs according to the prescribed tasks.

To add one-dof on the mechanism defined in section 3, we introduce one revolute joint between the previous mobile platform and the tool frame. Only one leg is necessary to hold the tool frame in position. Figure 4 shows the mechanism obtained with two translational dofs and one rotational dof.

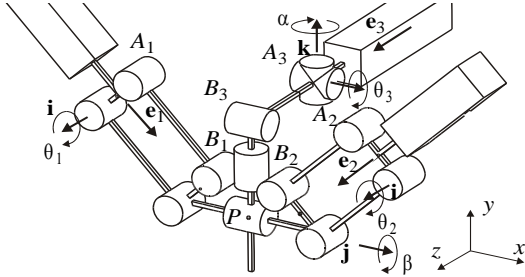

Figure 4: Parallel mechanism with two-dof of translation and one-dof of rotation

The architecture of the leg added is *PUU* where *P* and *U* stand for Prismatic and Universal joints, respectively [6]. The new prismatic joint is located orthogonaly to the first two prismatic joints. This location can be easily justified because on this configuration, *i.e.* when $\mathbf{b}_1 - \mathbf{a}_1 \perp \mathbf{b}_2 - \mathbf{a}_2$ and $\mathbf{b}_3 - \mathbf{a}_3 \perp \mathbf{b}_3 - \mathbf{p}$, the third leg is far away from serial and parallel singularities.

Let $\dot{\boldsymbol{\rho}}$ be referred to as the vector of actuated joint rates and $\dot{\mathbf{p}}$ as the velocity vector of point *P*,
$$\dot{\boldsymbol{\rho}} = [\dot{\rho}_1 \ \dot{\rho}_2 \ \dot{\rho}_2]^T \text{ and } \dot{\mathbf{p}} = [\dot{x} \ \dot{y}]^T$$

Due to the architecture of the two-dof mechanism and the location of *P*, its velocity on the z-axis is equal to zero. $\dot{\mathbf{p}}$ can be written in three different ways by traversing the three chains $A_i B_i P$,

$$\dot{\mathbf{p}} = \dot{\mathbf{a}}_1 + \dot{\theta}_1 \mathbf{i} \times (\mathbf{b}_1 - \mathbf{a}_1) \quad (3a)$$
$$\dot{\mathbf{p}} = \dot{\mathbf{a}}_2 + \dot{\theta}_2 \mathbf{i} \times (\mathbf{b}_2 - \mathbf{a}_2) \quad (3b)$$
$$\dot{\mathbf{p}} = \dot{\mathbf{a}}_3 + (\dot{\theta}_3 \mathbf{j} + \dot{\alpha} \mathbf{k}) \times (\mathbf{b}_3 - \mathbf{a}_3) + \dot{\beta} \mathbf{j} \times (\mathbf{p} - \mathbf{b}_3) \quad (3c)$$

where $\mathbf{a}_i$ and $\mathbf{b}_i$ are the position vectors of the points $A_i$ and $B_i$ for $i = 1, 2, 3$, respectively. Moreover, the velocities $\dot{\mathbf{a}}_1$, $\dot{\mathbf{a}}_2$ and $\dot{\mathbf{a}}_3$ of $A_1$, $A_2$ and $A_3$ are given by $\dot{\mathbf{a}}_1 = \mathbf{e}_1 \dot{\rho}_1$, $\dot{\mathbf{a}}_2 = \mathbf{e}_2 \dot{\rho}_2$ and $\dot{\mathbf{a}}_3 = \mathbf{e}_3 \dot{\rho}_3$, respectively.

We want to eliminate the passive joint rates $\dot{\theta}_i$ and $\dot{\alpha}$ from Eqs. (3a-c), which we do upon dot-multiplying Eqs. (3a-c) by $\mathbf{b}_i - \mathbf{a}_i$,

$$(\mathbf{b}_1 - \mathbf{a}_1)^T \dot{\mathbf{p}} = (\mathbf{b}_1 - \mathbf{a}_1)^T \mathbf{e}_1 \dot{\rho}_1 \quad (4a)$$
$$(\mathbf{b}_2 - \mathbf{a}_2)^T \dot{\mathbf{p}} = (\mathbf{b}_2 - \mathbf{a}_2)^T \mathbf{e}_2 \dot{\rho}_2 \quad (4b)$$
$$(\mathbf{b}_3 - \mathbf{a}_3)^T \dot{\mathbf{p}} = (\mathbf{b}_3 - \mathbf{a}_3)^T \mathbf{e}_3 \dot{\rho}_3 \\ + (\mathbf{b}_3 - \mathbf{a}_3)^T \dot{\beta} \mathbf{j} \times (\mathbf{p} - \mathbf{b}_3) \quad (4c)$$

Equations (4a-c) can be cast in vector form, namely,
$$\mathbf{t} = \mathbf{J} \dot{\boldsymbol{\rho}} \text{ with } \mathbf{J} = \mathbf{A}^{-1} \mathbf{B} \text{ and } \mathbf{t} = [\dot{x} \ \dot{y} \ \dot{\beta}]^T$$

where **A** and **B** are the parallel and serial Jacobian matrices, respectively,

$$\mathbf{A} \equiv \begin{bmatrix} (\mathbf{b}_1 - \mathbf{a}_1)^T & 0 \\ (\mathbf{b}_2 - \mathbf{a}_2)^T & 0 \\ (\mathbf{b}_3 - \mathbf{a}_3)^T & -(\mathbf{b}_3 - \mathbf{a}_3)^T \mathbf{j} \times (\mathbf{p} - \mathbf{b}_3) \end{bmatrix}$$

$$\mathbf{B} \equiv \begin{bmatrix} (\mathbf{b}_1 - \mathbf{a}_1)^T \mathbf{e}_1 & 0 & 0 \\ 0 & (\mathbf{b}_2 - \mathbf{a}_2)^T \mathbf{e}_2 & 0 \\ 0 & 0 & (\mathbf{b}_3 - \mathbf{a}_3)^T \mathbf{e}_3 \end{bmatrix}$$

There are two new singularities when one leg is added. The first one is a parallel singularity when $(\mathbf{b}_3 - \mathbf{a}_3)^T \mathbf{j} \times (\mathbf{p} - \mathbf{b}_3) = 0$, *i.e.*, when the lines $(A_3 B_3)$ and $(B_3 P)$ are colinear, and the second one is a serial singularity when $(\mathbf{b}_3 - \mathbf{a}_3)^T \mathbf{e}_3 = 0$, *i.e.*, $\mathbf{a}_3 - \mathbf{b}_3 \perp \mathbf{e}_3$. However, these singular configurations are simple and can be avoided by proper limits on the actuated joints.

## 3.3   Optimization of the useful workspace for translational motions

Two types of workspaces can be defined, (i) the Cartesian workspace is the manipulator's workspace defined in the Cartesian space, and (ii) the useful workspace is defined as a subset of the Cartesian workspace. Workspace and size are prescribed where some performance indices are prescribed.

For parallel mechanism, the useful workspace shape should be similar to the one of classical serial machine tools, which is parallelepipedic if the machine has three translational degrees of freedom for instance. So, a square useful workspace is prescribed here where the velocity amplification factors remain under the prescribed values. Two square useful workspaces can be used, (i) he first one has horizontal and vertical sides (Figure 5a) and (ii) the second one has oblique sides but its size is higher (Figure 5b).

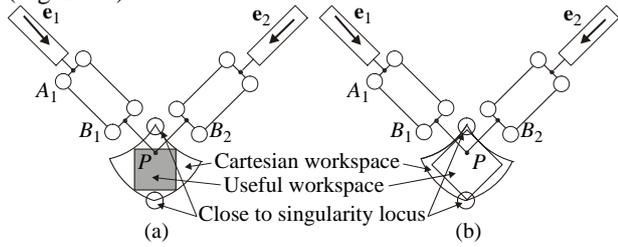

Figure 5: Cartesian workspace and isotropic configuration

To find the best useful workspace (center locus and size), we can shift the useful workspace along *x*-axis $(\Delta x)$ and *y*-axis $(\Delta y)$ (Figure 6) and the velocity amplification factors are computed for each configuration. This method was developed in [13].

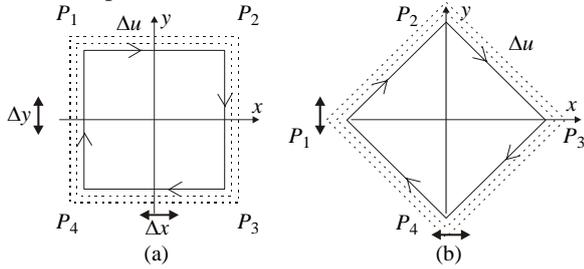

Figure 6: Looking for the best useful workspace (center locus and size)

In each case, velocity amplification factor extrema are located along the sides $P_i P_j$: they start from 1 at point S, then they vary until they reach prescribed boundaries ($1/3 \leq \lambda_i \leq 3$). Then computation (which analytical expressions $\lambda_i$) has been obtained with Maple along the four sides of the square and for the same length of the leg equal to one.

For the first mechanism, solution (a), the size of the optimal surface is equal to 0,89 m² and for the second mechanism, solution (b), the size is equal to 0,62 m². The result obtained for the solution (b) is smaller than for the solution (a) but is more appropriate for the extension three-axis mechanism of Figure 4. In effect, we want the axis of rotation to be parallel to one of the side of the useful workspace. In the next section, the lengths of the third leg will optimize to achieve this square useful workspace without singularity.

### 3.4 Optimization of the third-axis for rotational motions

The aim of this section is to define the two lengths of the leg, $L_1 = \|B_3 P\|$ and $L_2 = \|A_3 B_3\|$ such that it is possible to achieve the maximum range variation of third axis $\beta$ without meeting a singular configuration throughout the square workspace with the size defined in the previous section. The fist step of this optimization is to find the location of the prismatic joint. When *P* is on the center of the square workspace, we chose to place the third leg furthest away from singular configuration, *i.e.* when $\mathbf{b}_3 - \mathbf{a}_3$ and $\mathbf{e}_3$ are colinear and $\mathbf{b}_3 - \mathbf{a}_3 \perp \mathbf{p} - \mathbf{b}_3$ (Figure 7a).

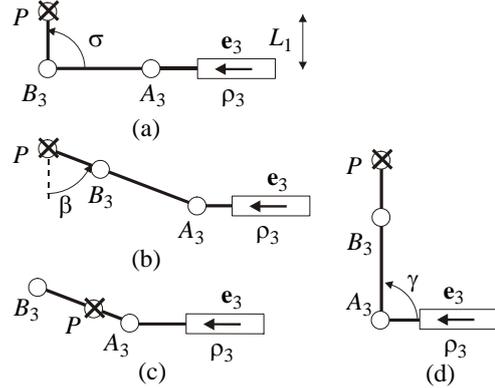

Figure 7: Optimal, serial and parallel configuration of the third leg

As it is defined in Section 3.2, the third leg is in a singular configuration whenever $(\mathbf{b}_3 - \mathbf{a}_3)^T \mathbf{j} \times (\mathbf{p} - \mathbf{b}_3) = 0$ (Figure 7b-c) or $(\mathbf{b}_3 - \mathbf{a}_3)^T \mathbf{e}_3 = 0$ (Figure 7d). In the optimization function, we set:

$$\frac{(\mathbf{b}_3 - \mathbf{a}_3)^T}{\|\mathbf{b}_3 - \mathbf{a}_3\|} \mathbf{e}_3 > 0,2 \qquad (4a)$$

$$\frac{(\mathbf{b}_3 - \mathbf{a}_3)^T}{\|\mathbf{b}_3 - \mathbf{a}_3\|} \mathbf{j} \times \frac{(\mathbf{p} - \mathbf{b}_3)}{\|\mathbf{p} - \mathbf{b}_3\|} > 0,2 \qquad (4b)$$

with $\arcsin(0,2) = 11,53°$

This means that $\gamma = \angle A_3 B_3 P$ is in $[11,5\ 168,5]$ and $\sigma = \angle A_3 B_3 \mathbf{e}_3$ is in $[-78,5\ 78,5]$. The result of this optimization as a function of $L_1$ and $L_2$ is depicted in Fig. 8.

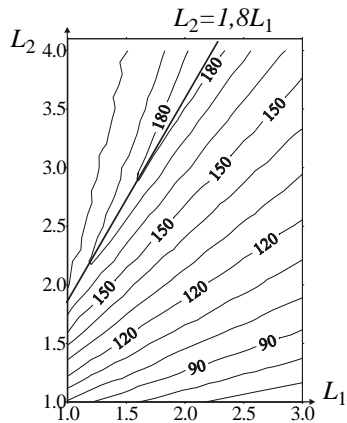

Figure 8: Range of variation in degrees of $\beta$ as a function of $L_1$ and $L_2$

With a suitable chose of lengths, it is easy to obtain a range variation of $\beta$ higher than 180°. So, this value can be reduced if we increase the constraint defined in Eqs. (4). However, when $L_2 = 1,8 L_1$, the range of variation is optimal.

## 4  Conclusions

In this paper, a parallel mechanism with two degrees of position and one degree of rotation is studied. All the actuated joints are fixed prismatic joints, which can be actuated by means of linear motors or by conventional rotary motors with ball screws. Only three types of joints are used, *i.e.*, prismatic, revolute and universal joints. All the singularities are characterized easily because position and orientation are decoupled for the direct kinematic problem and can be avoided by proper design. The lengths of the legs as well as their positions is optimized, to take into account the velocity amplification factors for the translational motions and to avoid the singular configuration for the rotational motions.

## 5  Acknowledgments

This research was partially supported by the CNRS (Project ROBEA "Machine à Architecture compleXe").